\documentclass[10pt,twocolumn,letterpaper]{article}

\usepackage[pagenumbers]{cvpr} 

\usepackage[dvipsnames]{xcolor}

\definecolor{cvprblue}{rgb}{0.21,0.49,0.74}
\usepackage[pagebackref,breaklinks,colorlinks,citecolor=cvprblue]{hyperref}

\usepackage{booktabs} 
\usepackage{multirow}
\usepackage{makecell} 
\usepackage{comment}
\usepackage{appendix}
\usepackage{amsmath}
\usepackage{amssymb}
\usepackage{subcaption}
\usepackage{float}
\usepackage{afterpage}
\usepackage{caption}
\usepackage{placeins}
\usepackage[utf8]{inputenc}
\usepackage{amsfonts}

\usepackage{mathrsfs}

\def\paperID{12602} 
\def\confName{CVPR}
\def\confYear{2024}

\title{AdCorDA: Classifier Refinement via Adversarial Correction and Domain Adaptation}

\author{Lulan Shen, Ali Edalati, Brett Meyer, Warren Gross, James J. Clark\\
McGill University\\
Montreal, Quebec, Canada\\
{\tt\small lulan.shen@mail.mcgill.ca, james.j.clark@mcgill.ca}
}

\begin{document}
\maketitle
\begin{abstract}
This paper describes a simple yet effective technique for refining a pretrained classifier network. The proposed \mbox{AdCorDA} method is based on modification of the training set and making use of the duality between network weights and layer inputs. We call this input space training. The method consists of two stages - adversarial correction followed by domain adaptation. Adversarial correction uses adversarial attacks to correct incorrect training-set classifications. The incorrectly classified samples of the training set are removed and replaced with the adversarially corrected samples to form a new training set, and then, in the second stage, domain adaptation is performed back to the original training set. Extensive experimental validations show significant accuracy boosts of over 5\% on the CIFAR-100 dataset. The technique can be straightforwardly applied to refinement of weight-quantized neural networks, where experiments show substantial enhancement in performance over the baseline.
The adversarial correction technique also results in enhanced robustness to adversarial attacks.
\end{abstract}    
\section{Introduction - Input Space Training}
\label{sec:intro}



In this paper, we present an alternative to standard neural network training methods. Rather than modifying network weights driven by errors relative to a training dataset, we consider \emph{input space training}, which is driven by the effect of changing the inputs on the loss, rather than (only) the effect of changing the network weights on the loss.
As noted by Feng and Tu \cite{feng2022activityweight}, there is a duality between neural network layer inputs (activations) and weights with respect to the loss function $L$, due to the mathematical form of the standard single-layer perceptron - $y = f(w^Tx)$. For a given change in the loss function value due to a small change in the weights $w$, one can get an equivalent change in the loss by changing the input activations, $x$, instead. That is, $\Delta L(x_0,w_0) = L(x_0+\delta x,w_0)-L(x_0,w_0) = L(x_0,w_0+\delta w)-L(x_0,w_0)$. 
This observation of activity-weight duality leads to the following learning procedure, which we call \emph{input space training} (IST), based on manipulation of the inputs:
\begin{enumerate}
    \item Make a small change in the inputs $x$ that results in a reduction of the loss. 
    \item Use the duality between weights and activations to determine a change in the weights that gives a reduction in the loss for the original inputs. 
\end{enumerate}
To carry out this procedure, we need to accomplish two tasks - find a change in inputs that reduces the loss, and find an equivalent change in weights that reduces the loss on the original inputs. The paper by Feng and Tu \cite{feng2022activityweight} provides a method for accomplishing the second task. They note that there are, in general, more weight parameters than inputs, and so there are effectively unlimited possible weight changes that are equivalent to a given change in the inputs. They provide a minimum weight change solution to identify a unique weight update, as a linear combination of the current weight values.

Feng and Tu use the activity-weight duality principle for the purposes of quantifying how networks generalize, but do not use it to train a network. One could derive a training scheme from their equations but it would be slow and computationally expensive. In this paper, rather than training a network from scratch using input space training, we propose a method which takes a classifier network pretrained using standard back-propagation methods and then refines it using a one-shot (non-iterative) IST method. Inspired by activity-weight duality, our IST refinement method has two stages: first we perturb the training set to reduce the loss, and second, we adjust the weights by performing a domain adaptation step that adapts from the perturbed dataset to the original dataset.

\section{Curriculum Learning}
\label{sec:curriculum learning}

The first approach we propose for altering the training set to reduce loss is based on \emph{curriculum learning}. Curriculum learning, first proposed by Bengio et. al \cite{Bengio_curriculumlearning}, aims to improve the speed and accuracy of network training, by presenting data samples from the training set in an ordered fashion. Typically, \emph{easier} samples are presented before \emph{difficult} samples as the training progresses.

It is not obvious how to properly define the notions of ``easy" and ``hard", however, and indeed many different definitions exist. Some of these definitions are based solely on the structure of the input examples, without consideration of the network being trained. Table 2 in the survey paper of Wang et. al \cite{wang2021survey} lists no less than nineteen different types of pre-defined input difficulty measures that have been used to guide curriculum learning. But the difficulty of an input can also depend on the network being trained. Problems that some networks find difficult may be easy for other networks, and vice-versa. So-called \emph{Self-Paced-Learning (SPL)} methods, such as proposed by Kumar et. al \cite{kumar2010self} use dynamic measures of problem difficulty that are provided by the network itself as it trains. In the SPL method, easy problems are defined as those problems for which the network's training loss is less than a (dynamically changing) threshold value.

We propose to use the curriculum separation of the training set into easy and hard problems, as defined by the training loss threshold, for our IST approach. We consider that, over the original training set, our pre-trained network achieves a particular loss value. If we \emph{remove} the training set samples for which the loss is above a threshold, then we are left with a (modified) training set for which the average (and maximum) loss is less than that of the original training set. To avoid having to set a suitable threshold value, we propose using the pre-trained network to define easy vs. hard using the simple expedient of considering easy problems to be ones the network classifies correctly. This will naturally result in a separation of input samples based on loss. We use this procedure to satisfy the first step of our IST process - altering the inputs to reduce the loss. Even though we are not altering individual samples in this method, the collective of the samples on a batch level \emph{is} being altered.

\section{Adversarial Correction}
\label{sec:adversarial correction}

We can take the curriculum approach outlined in the previous section a step further, by doing what we call \emph{adversarial correction} to further modify the training set. This results in a larger set of inputs for which the loss is reduced than the curriculum approach. 

The concept of adversarial attack is well known in the machine learning community \cite{Li2022ReviewAdversarial}. Given a classifier network trained on a particular dataset, an adversary can modify an input slightly in such a way that the network gives a different classification output. One thing to keep in mind, however, especially for smaller networks, is that networks frequently give wrong answers, even in the absence of adversarial attacks. For example, as seen in \cref{table:resnet_fp_cifar_3seed}, the baseline accuracy of small ResNet networks on the CIFAR-100 dataset \cite{Krizhevsky2009Learning} is around 80\%, meaning that, even without considering possible adversarial attacks, such networks give the wrong output 20\% of the time on the CIFAR-100 test set. However, because of this relatively frequent failure, in the application of such networks, especially on edge devices or embedded systems, network outputs are often combined with measures of uncertainty or confidence. This allows the user to judge whether a particular output can be trusted. Although this check is not perfect, it does mitigate the impact of adversarial attacks. In addition to that, there have been many training methods \cite{Mary2018Towards, Shafahi2019Adversarial} developed that increase the robustness of networks to adversarial attacks.

In this paper, rather than focusing on correct outputs being changed by adversarial attacks, we look at the effect of adversarial attacks on the outputs that the network already gets wrong. In such a situation, things cannot get any worse, as the network is already wrong, but they could get better if the adversarial perturbation of the input actually causes the network to provide the \emph{correct} answer. We can help the process by using \emph{targeted} attacks, where the target of the adversarial attack in this case is the correct output. But even non-targeted attacks may help by weakening support for the incorrect label relative to the true label. We will refer to this as \emph{adversarial correction}, as opposed to \emph{adversarial attack}.

It should be noted that adversarial correction is well-suited to working with quantized networks, as some adversarial attacks do not need to compute the gradients with respect to the weights. However, many attacks do need gradient information and deep domain adaptation techniques generally require gradient-based optimization (gradients with respect to the weights) to adapt models effectively across domains. Thus, in this paper, we focus on post-training quantization methods \cite{Jacob2018Quantization}, and we apply the adversarial correction on the samples the quantized network gets wrong, rather than those of the full precision network. 
\section{Domain Adaptation}
\label{sec:domain adaptation}

At this point in the method we have a modified dataset consisting of either only samples that the original network gets correct, or the same augmented with samples that have been adversarially corrected. Either way, our original trained network has an accuracy of $100\%$ on this modified dataset. But, how does this help us? After all, what we really want to do is increase accuracy (reduce the loss) on the original dataset, not some other dataset. This is the goal of the second stage of the input space training, namely finding a set of network weights that results in a lower loss on the original training set, starting from the modified training set.

Denote the original training set by $T$, and consider the altered training set $T^\prime$ as our starting point for the second stage of the IST process. The original training set can be thought of as a distribution shift of the altered training set. How can we deal with this distribution shift, where we go from a distribution where the network does well (perfectly, in fact), to a distribution where the network performs less well? There is substantial literature addressing this very problem: \emph{domain adaptation}. Domain adaptation methods aim to transfer knowledge about one domain (the \emph{source} domain) into a second, similar, domain (the \emph{target} domain) \cite{zhang2021survey}. All domain adaptation methods have the goal of increasing performance on the target domain, starting from a network that does well on the source domain. 
Shen et. al \cite{Shen2023Fast} showed that applying domain adaptation from easy to hard after the early stages of curriculum learning speeds up training. 
Motivated by these considerations we choose the final step in our IST method to be a domain adaptation from $T^\prime$ to $T$.

\section{AdCorDA}
\label{sec:AdCorDA}

\begin{figure}[hbt!]
        \centering
        \includegraphics[width=0.45\textwidth]{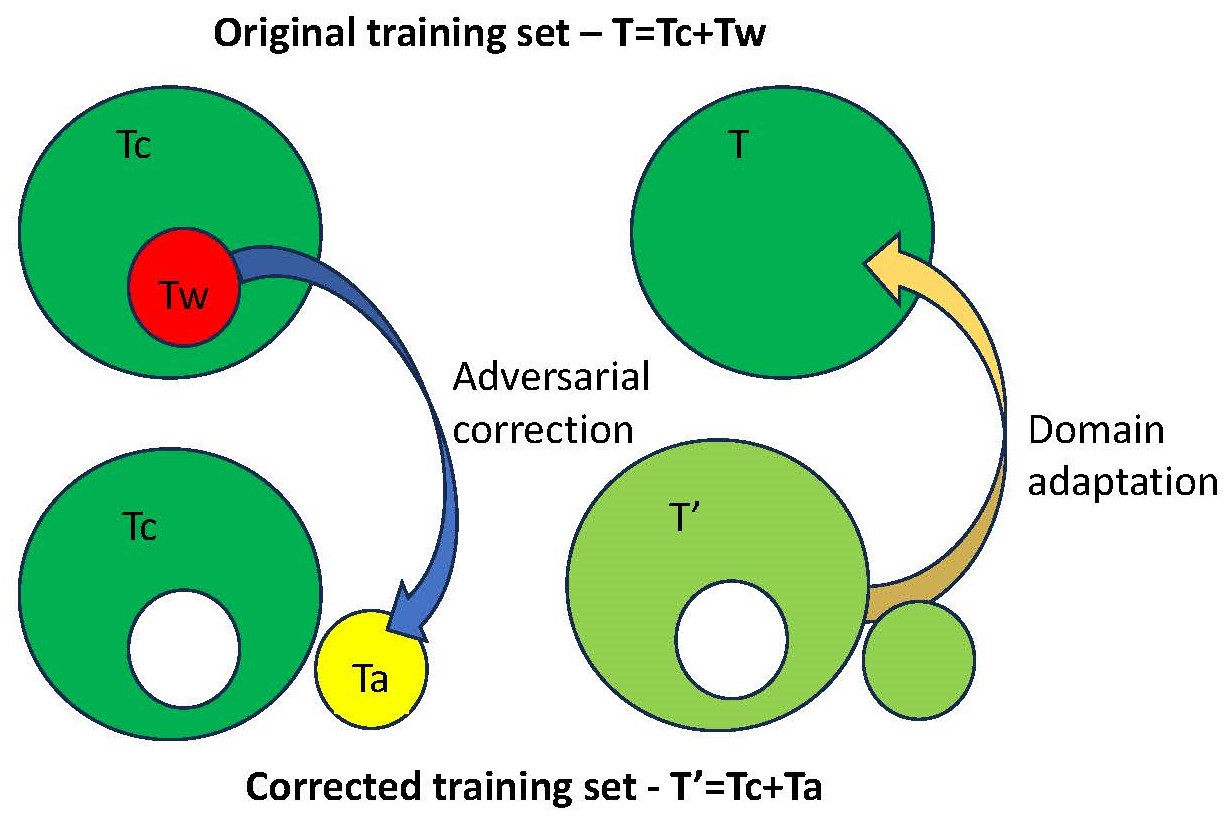}
        \caption{Overview of the proposed AdCorDA classifier refinement method. $T$ is the original training set; $T_c$ is the subset of $T$ that the pretrained network labels correctly, and $T_w$ the subset that is labeled incorrectly; $T_a$ is the set of samples that have been adversarially corrected; $T^\prime$ is the union of $T_c$ and $T_a$. The network is adapted from $T^\prime$ as the source domain back to $T$ as the target domain.}
    \label{overview}
\end{figure}

Putting together the two stages of the input space training method as detailed above, we arrive at what we call the AdCorDA (Adversarial Correction and Domain Adaptation) method. The AdCorDA method proceeds as depicted in \cref{overview}, with the following steps:

\begin{itemize}
\item Step 1: train a network to solve a classification problem using standard training techniques on a training set $T$.
\item Step 2: separate the original set of training samples $T$ into two subsets: $T_c$ and $T_w$, where $T_c$ are training samples for which the trained network gets the correct answer, and $T_w$ are training samples for which the network gets wrong.
\item
Step 3: for each sample in $T_w$, use adversarial attack techniques to create adversarial inputs, where in this case we wish to perturb the input such that the network gives the class provided by the training label (true label). Note that typically not all attacks will successfully coax the network to output the true label. Let the set of successfully perturbed samples be denoted as $T_a$. This may be smaller than the set $T_w$.
\item Step 4: Merge the subsets $T_c$ and $T_a$ into one new training set, $T^\prime$. The samples for which the adversarial correction failed have been removed, so the accuracy of the network on $T^\prime$ is $100\%$, and the number of elements in $T^\prime$ may be less than that of the original dataset $T$.
\item Step 5: Seeing that $T$ and $T^\prime$ represent two (overlapping) domains, do \emph{domain adaptation} of the trained network, adapting from the corrected dataset $T^\prime$ as the source domain, back to the original dataset $T$ as the target domain.
\end{itemize}

In the experiments described in the next section, we examine the effectiveness of the AdCorDA method, as well as an ablation case where we omit steps 3 and 4, using only the curriculum subset as $T^\prime$. 
\section{Experimental Setup}
\label{sec:experimental setup}

\subsection{Datasets, Networks and Training Details}
We validated our approach through experiments on the CIFAR-10 and -100 datasets, each containing 50K images, which are randomly split into 45K training data and 5K validation data. Each dataset has a separate test set of 10K images. We first initialize ResNets \cite{he2015deep} of different sizes (i.e., ResNet-18, ResNet-34, ResNet-50) and EfficientNetV2-M \cite{Tan2021EfficientNetV2} with parameters pre-trained on the ImageNet dataset \cite{ImageNet} from PyTorch \cite{paszke_pytorch_2019} and then fine-tune \cite{yosinski2014transferable} on the CIFAR training sets to obtain the corresponding baseline models. Input images are resized to $224 \times 224$ and use the same data transform (i.e., apply normalization using the mean and standard deviation of ImageNet data) as the pre-trained models. During the fine-tuning, we use a stochastic gradient descent (SGD) optimizer \cite{SGD} with a momentum of 0.9, a weight decay of 1e-4, a batch size of 128 for ResNets on and of 64 (due to limitations in computing resources) for EfficientNetV2-M, a fixed learning rate of 1e-4, and we train for a total of 100 epochs on both CIFAR datasets. 
We define the fine-tuned models with the best validation accuracy as our baseline models. 

We use PyTorch and Nvidia V100L and P100L GPUs for all implementations. We split the training and validation datasets using three random seeds: 1, 2, and 5. We find that using random seeds 3 and 4 does not allow some attacks (e.g., BI and BIH, described in the next subsection) to successfully attack any incorrect images in the majority of instances. Therefore, we present the averaged results of the main experiments based on the three chosen random seeds.
To determine the optimal hyper-parameters for our model, we perform a basic parameter grid search for the batch size, base learning rate, and weight decay of the SGD optimizer.

\subsection{Adversarial Attack Methods}
To apply adversarial attacks on misclassified images of train domains, we use a selection of methods, including three major types of gradient-based attacks: basic iterative method \cite{Kurakin2017Adversarial} and its variants, iterative least likely class \cite{Kurakin2017Adversarial}, decoupled direction and norm \cite{rony2019decoupling}, as well as a non-gradient-based salt and pepper noise attack, briefly described below.

\begin{itemize}
    \item \textbf{Untargeted Basic Iterative (BI)} \cite{Kurakin2017Adversarial}: This extends the ``fast" method \cite{Goodfellow2015Explaining}, which generates adversarial images through iterative processes using a small step size ($\alpha$) and clip pixel values of intermediate results at each step to ensure that they remain within an $\epsilon$-neighbourhood of the source image \cite{Kurakin2017Adversarial}:
    \begin{align}
        \boldsymbol{X}_{N+1}^{BI} &= Clip_{X, \epsilon} \{ \boldsymbol{X}_{N}^{BI} + \alpha \text{sign}(\nabla_X J (\boldsymbol{X}_{N}^{BI}, y_{true})) \},\\
        \boldsymbol{X}_0^{BI} &= \boldsymbol{X}, 
        \label{eqn:bi} 
    \end{align}
    where $\boldsymbol{X}$ represents an image, $y_{true}$ denotes the true class for the image $\boldsymbol{X}$, $J (\boldsymbol{X}, y)$ is the cross-entropy cost function of the neural network, $Clip_{X, \epsilon} \{ \boldsymbol{X'}\}$ is the per-pixel clipping function applied to the image $\boldsymbol{X'}$ to ensure it falls within an $L_\infty$ $\epsilon$-neighbourhood of the original image $\boldsymbol{X}$.
    
    \item \textbf{Basic Iterative method with Highest probability class (BIH)}: When attacking a correct image, BI uses the true class gradient, where the highest probability class aligns with the true class. However, when targeting an incorrect output, this changes – the highest probability class no longer represents the truth. Therefore, we adapt BI to use the gradient of the highest probability class to weaken the accuracy of the incorrect output, illustrated below:  
    \begin{align}
    \boldsymbol{X}_{N+1}^{BIH} &= Clip_{X, \epsilon} \{ \boldsymbol{X}_{N}^{BIH} + \alpha \text{sign}(\nabla_X J (\boldsymbol{X}_{N}^{BIH}, y_{H})) \}\label{eqn:bih_1} \\
        y_H &= \underset{y}{\mathrm{argmax}} \{p(y|\boldsymbol{X})\}.\label{eqn:bih_2}
    \end{align}
    \normalsize 
    
    \item \textbf{Targeted Variant of Basic Iterative (VBI)}: In addition to the standard untargeted BI method, we created a targeted variant called VBI. Unlike BI (Eq.~\eqref{eqn:bi}), which moves away from the true label, VBI (Eq.~\eqref{eqn:vbi}) operates in the opposite direction, moving towards the true label by negating the sign of the gradient sign function. 
    \begin{equation}
        \boldsymbol{X}_{N+1}^{VBI} = Clip_{X, \epsilon} \{ \boldsymbol{X}_{N}^{VBI} - \alpha \text{sign}(\nabla_X J (\boldsymbol{X}_{N}^{VBI}, y_{true})) \} \label{eqn:vbi} 
    \end{equation}
    $VBI_{iter1}$ is a fast version of VBI that just does one step towards the target.
    \item \textbf{Iterative Least-Likely class (LL)} \cite{Kurakin2017Adversarial}: This method generates an attack targeting the least-likely class, as predicted by the trained model on the source image:
    \begin{align}
    \boldsymbol{X}_{N+1}^{LL} &= Clip_{X, \epsilon} \{ \boldsymbol{X}_{N}^{LL} - \alpha \text{sign}(\nabla_X J (\boldsymbol{X}_{N}^{LL}, y_{LL})) \}\label{eqn:LL_1} \\
        y_{LL} &= \underset{y}{\mathrm{argmin}} \{p(y|\boldsymbol{X})\} \label{eqn:LL_2}
    \end{align}
    \normalsize 
    The LL method moves the input in the direction of the gradient toward the least probable class. While this may lower the probability of the true class, in some cases it will also lower the probability of the maximum probability (incorrect) class by a larger amount, potentially resulting in a correction of the output label.
    \item \textbf{Decoupled Direction and Norm (DDN)} \cite{rony2019decoupling}: This attack is an iterative approach that refines the noise added to the input image in each iteration to make it adversarial.
    At iteration $i$, the adversarial input image, $x_i$, is generated as $x_i=x+\eta_i$, where $\eta_i$ is the noise with a norm of $\sigma_i$. If $x_i$ is adversarial, the norm of the next iteration noise is decreased ($\sigma_{i+1}=\sigma_i(1-\epsilon)$). Otherwise, the norm of the next noise is increased ($\sigma_{i+1}=\sigma_i(1+\epsilon)$). This process repeats until the minimum required perturbation is found \cite{rony2019decoupling}. The DDN method is a targeted attack that moves the network output towards the true label.
    
    \item \textbf{Salt and Pepper noise (SP)}: A non-gradient-based attack that repeatedly adds Salt \& Pepper noise to the input to fool the model.
\end{itemize}

For the DDN and SP attacks, we use the default hyper-parameters provided by the Foolbox
framework \cite{rauber2017foolbox,rauber2017foolboxnative}. Note that the input images are subject to the ImageNet transformation with a lower and upper bound of 0 and 1, respectively. The BI and LL attacks are applied according to the experimental setting outlined in \cite{Kurakin2017Adversarial}. In \cref{table:resnet_fp_cifar_3seed}, we have applied VBI with one iteration (referred to as VBI$_\text{iter1}$) and with five iterations (referred to as VBI). The maximum iteration limit for VBI is set to 5, which is sufficient for effectively correcting the vast majority of erroneous samples on CIFAR datasets. 

We desire adversarial attack methods which are fast. The adversarial correction process may be quite computationally expensive if many iterations are needed per element of $T_w$. For example, a small network trained on CIFAR-100, which has 50,000 samples in the training set, might have a training accuracy of 95\%. In this case, there would be 2,500 incorrect samples needing to be adversarially corrected. This could take a lot of time if the method takes many expensive iterations. Thus, we would prefer adversarial attack methods which are quick, using only a few iterations, possibly at the cost of not being able to correct all samples. 
Note that we do not necessarily require that the corrections be imperceptible, and so we can use relatively large perturbations of the input.

To investigate the effect of our proposed method on the adversarial robustness of the corrected models, we evaluated the models against \textit{AutoAttack}
\cite{AutoAttack} on CIFAR-10 and CIFAR-100 test sets. Composed of four different attacks from those used in our experiments for the correction, AutoAttack is a well-known, powerful, and diverse ensemble of parameter-free attacks. We applied the standard version of AutoAttack: $\text{APGD}_{\text{CE}}$, targeted $\text{APGD}_{\text{DLR}}$ \cite{AutoAttack}, targeted FAB \cite{FAB}, and Square Attack \cite{SquareAttack} with $\ell_{\infty}$-norm. The attacks were applied sequentially. We set $\epsilon$ to 5e-4 for all of the AutoAttack experiments. Other AutoAttack parameters such as iterations and number of restarts are identical to the parameters used in the standard version. The batch size used for ResNet-18, ResNet-34, and EfficientNetV2-M experiments is 512, 512, and 100, respectively. We reported the average accuracy obtained across three different random seeds: 1, 2, and 5.

\subsection{Domain Adaptation Method}
In the domain adaptation stage, we utilize Deep CORAL \cite{sun2016deep}, which aligns the second-order covariance matrices between a source domain and a target domain through CORAL loss. This alignment helps to bridge the distribution gap between the domains and improve the model's performance on the target domain. Aligning the implementation with the original paper, CORAL loss is only applied to the last classification layer in the neural networks. The total loss is the sum of the classification loss and the CORAL loss, defined as
\begin{equation}
    \mathcal{L}_{loss} = \mathcal{L}_{class} + \lambda \mathcal{L}_{coral},
    \label{eqn:coral loss}
\end{equation}
where $\lambda$ is a weight between classification and CORAL loss. Its value is 1/750 for CIFAR-10 and 1/25 for CIFAR-100, intending to align the classification loss and the CORAL loss nearly the same at the end of the training process. The CORAL loss term is given by the following equation \cite{sun2016deep}:
\begin{equation}
    \mathcal{L}_{coral} = \frac{1}{4d^2} ||C_S-C_T||^2_F,
\end{equation}
where $C_S$ and $C_T$ are the covariance matrices of features induced by samples from the source domain and target domain, respectively, and the norm is the squared-matrix Frobenius norm.
In our application, the source domain is the adversarially corrected training dataset ($T^\prime$) and the target domain is the original training dataset ($T$). 

Our experimental setup closely adheres to the guidelines in \cite{sun2016deep}. However, we deviate by using batch sizes of 16 for ResNets on CIFAR-10 and CIFAR-100 and 16/32 for EfficientNet-M on CIFAR-10/100, differing from the original paper's settings. 
Note that we shuffle the dataset $T^\prime$ before conducting domain adaptation training. This ensures a mixture of training samples, including those from the original dataset for which the trained network gets the correct answers and the successfully perturbed incorrect samples. We then proceed with 20 epochs on the CIFAR datasets. We also initialize the model with pre-trained weights from the baseline models rather than using the pre-trained model on ImageNet from PyTorch. These adjustments ensure a fair comparison with baseline models. When applying domain adaptation to quantized models, we facilitate the back-propagation process by approximating the gradients in these models. We achieve this approximation by utilizing the gradients derived from their corresponding full-precision models. This approach enables us to effectively conduct back-propagation on the quantized models. We define the best adapted model as the one that achieves the highest validation accuracy on the target domain, the original dataset $T$. 

\subsection{Network Quantization Method}
We also test the effectiveness of the AdCorDA method on network quantization, which reduces the precision of computations and weight storage by using lower bit-widths instead of floating-point precision. In our experiments, we choose post-training static quantization (PTSQ) \cite{Jacob2018Quantization}, which is one of the most common and fastest quantization techniques in practice. This technique determines the scales and zero points prior to inference. Specifically, we quantize the full-precision 32-bit (FP32) weights (\eg, $w \in [\alpha, \beta]$) and activations of the trained baseline models to 8-bit integer (Int8) values (\eg, $w_q \in [\alpha_q, \beta_q]$). The quantization process is defined as
\begin{equation}
    w_q = \text{round} \left(\frac{1}{s} w + z \right),
\end{equation}
where $s$ is the scale, and $d$ is the zero-point, defined as
\begin{equation}
    s = \frac{\beta-\alpha}{\beta_q - \alpha_q}, \quad z = \text{round} \left(\frac{\beta \alpha_q - \alpha \beta_q}{\beta - \alpha} \right).
\end{equation}

To obtain quantized models, we compress the baseline models using post-training static quantization (PTSQ) \cite{Jacob2018Quantization}. We use the built-in quantization modules provided by PyTorch. These modules facilitate the fusion of different model components, calibration of the model using training data to determine suitable scale factors, and the actual quantization of weights and activations in the model. Note that we perform the adversarial correction on the training samples that the quantized network gets wrong, not the ones that the full precision network gets wrong.
\section{Results and Discussion}
\label{sec:results and discussion}

\begin{table*}[hbt!]
\centering
\resizebox{\textwidth}{!}{%
\begin{tabular}{ccccccccccccccc} \toprule
    \multirow{2.5}{*}{Model} & \multirow{2.5}{*}{Approach} & \multirow{2.5}{*}{Attack} & \multicolumn{6}{c}{CIFAR-10} & \multicolumn{6}{c}{CIFAR-100} \\ \cmidrule(lr){4-9} \cmidrule(lr){10-15} 
    & & & Corr. rate & $T^\prime$ & Train & Valid & Test & $\Delta$ Acc & Corr. rate & $T^\prime$ & Train & Valid & Test & $\Delta$ Acc\\\midrule
    \multirow{8}{*}{\makecell{ResNet-18 \\(11.19M)}} & BL & - & - & - & 99.61 $\pm$ 0.56 & 93.73 $\pm$ 0.43 & 93.29 $\pm$ 0.37 & - & - & - & 99.00 $\pm$ 1.19 & 76.84 $\pm$ 0.12 & 77.04 $\pm$ 0.08 & -\\
    & BL-IST & None & - & 99.96 $\pm$ 0.04 & 99.69 $\pm$ 0.36 & 95.91 $\pm$ 0.20 & 95.57 $\pm$ 0.13 & +2.28 & - & 98.86 $\pm$ 0.94 & 98.84 $\pm$ 0.98 & 80.14 $\pm$ 0.47 & 80.27 $\pm$ 0.74 & +3.23\\
    & BL-IST & LL & 55/176 & 100.00 & 99.80 $\pm$ 0.28 & 96.17 $\pm$ 0.08 & 95.93 $\pm$ 0.15 & +2.64 & 70/451 & 100.00 & 99.20 $\pm$ 0.92 & 80.99 $\pm$ 0.21 & 80.93 $\pm$ 0.46 & +3.90\\
    & BL-IST & BIH & 99/176 & 100.00 & 99.86 $\pm$ 0.19 & 96.16 $\pm$ 0.28 & 95.87 $\pm$ 0.24 & +2.58 & 51/451 & 100.00 & 99.36 $\pm$ 0.72 & 80.75 $\pm$ 0.54 & \textbf{80.99 $\pm$ 0.45} & \textbf{+3.96}\\
    & BL-IST & VBI$_\text{iter1}$ & 121/176 & 100.00 & 99.59 $\pm$ 0.46 & 96.09 $\pm$ 0.22 & \textbf{95.97 $\pm$ 0.12} & \textbf{+2.68} & 226/451 & 100.00 & 99.47 $\pm$ 0.59 & 80.81 $\pm$ 0.18 & 80.92 $\pm$ 0.56 & +3.89\\
    & BL-IST & VBI & 175/176 & 100.00 & 99.97 $\pm$ 0.04 & 96.19 $\pm$ 0.15 & 95.77 $\pm$ 0.06 & +2.48 & 446/251 & 100.00 & 99.80 $\pm$ 0.20 & 80.37 $\pm$ 0.98 & 80.54 $\pm$ 0.80 & +3.50\\
    & BL-IST & DDN & 176/176 & 100.00 & 100.00 & 96.21 $\pm$ 0.28 & 95.84 $\pm$ 0.07 & +2.55 & 451/451 & 99.98 $\pm$ 0.01 & 99.98 $\pm$ 0.01 & 80.79 $\pm$ 0.45 & 80.82 $\pm$ 0.35 & +3.79\\
    & BL-IST & SP & 45/176 & 100.00 & 99.79 $\pm$ 0.29 & 96.17 $\pm$ 0.12 & 95.80 $\pm$ 0.08 & +2.51 & 43/251 & 100.00 & 99.16 $\pm$ 1.00 & 80.63 $\pm$ 0.54 & 80.89 $\pm$ 0.61 & +3.86\\
    \midrule
    \multirow{8}{*}{\makecell{ResNet-34 \\(21.30M)}} 
    & BL & - & - & - & 99.43 $\pm$ 0.67 & 94.71 $\pm$ 0.05 & 94.22 $\pm$ 0.06 & - & - & - & 94.36 $\pm$ 2.24 & 78.12 $\pm$ 0.79 & 78.41 $\pm$ 0.10 & -\\ 
    & BL-IST & None & - & 99.92 $\pm$ 0.03 & 99.81 $\pm$ 0.10 & 96.78 $\pm$ 0.08 & 96.40 $\pm$ 0.05 & +2.18 & - & 95.38 $\pm$ 1.56 & 95.26 $\pm$ 1.64 & 82.99 $\pm$ 0.48 & 82.98 $\pm$ 0.07 & +4.57 \\
    & BL-IST & LL & 25/80 & 99.98 $\pm$ 0.02 & 99.89 $\pm$ 0.07 & 96.53 $\pm$ 0.16 & 96.31 $\pm$ 0.12 & +2.09 & 370/2538 & 100.00 & 96.05 $\pm$ 1.39 & 83.13 $\pm$ 0.08 & 82.69 $\pm$ 0.12 & +4.28\\
    & BL-IST & BIH & 46/80 & 99.99 $\pm$ 0.01 & 99.94 $\pm$ 0.06 & 96.53 $\pm$ 0.23 & 96.36 $\pm$ 0.07 & +2.14 & 655/2538 & 100.00 & 97.31 $\pm$ 1.19 & 83.04 $\pm$ 1.19 & 83.31 $\pm$ 0.06 & +4.90\\
    & BL-IST & VBI$_\text{iter1}$ & 53/80 & 99.99 & 99.97 $\pm$ 0.01 & 96.62 $\pm$ 0.07 & 96.26 $\pm$ 0.12 & +2.04 & 1207/2538 & 99.99 & 97.40 $\pm$ 0.92 & 83.39 $\pm$ 0.44 & 83.11 $\pm$ 0.23 & +4.70\\
    & BL-IST & VBI & 80/80 & 100.00 & 100.00 & 96.71 $\pm$ 0.22 & 96.26 $\pm$ 0.12 & +2.04 & 2490/2538 & 100.00 & 99.21 $\pm$ 0.12 & 83.34 $\pm$ 0.36 & 83.26 $\pm$ 0.45 & +4.85\\
    & BL-IST & DDN & 80/80 & 100.00 & 100.00 & 96.71 $\pm$ 0.22 & \textbf{96.71 $\pm$ 0.05} & \textbf{+2.49} & 2538/2538 & 99.98 $\pm$ 0.01 & 99.97 $\pm$ 0.01 & 83.55 $\pm$ 0.53 & \textbf{83.64 $\pm$ 0.06} & \textbf{+5.23}\\
    & BL-IST & SP & 23/80 & 99.98 $\pm$ 0.01 & 99.90 $\pm$ 0.09 & 96.52 $\pm$ 0.12 & 96.22 $\pm$ 0.05 & +2.00 & 118/2538 & 100.00 & 95.74 $\pm$ 1.48 & 83.33 $\pm$ 0.37 & 83.25 $\pm$ 0.29 & +4.84\\
    \midrule
    \multirow{8}{*}{\makecell{ResNet-50 \\ (23.57M)}} & BL & - & - & - & 99.81 $\pm$ 0.14 & 95.36 $\pm$ 0.36 & 94.32 $\pm$ 0.59 & - & - & - & 98.81 $\pm$ 0.73 & 80.01 $\pm$ 0.65 & 79.74 $\pm$ 0.19 & -\\
    & BL-IST & None & - & 99.92 $\pm$ 0.03 & 99.78 $\pm$ 0.04 & 96.65 $\pm$ 0.16 & \textbf{96.61 $\pm$ 0.12} & \textbf{+2.29} & - & 99.84 $\pm$ 0.01 & 98.34 $\pm$ 0.38 & 83.70 $\pm$ 0.14 & \textbf{83.89 $\pm$ 0.22} & \textbf{+4.15} \\
    & BL-IST & LL & 46/131 & 99.96 $\pm$ 0.01 & 99.84 $\pm$ 0.01 & 96.57 $\pm$ 0.19 & 96.31 $\pm$ 0.11 & +1.99 & 60/775 & 99.99 $\pm$ 0.01 & 98.58 $\pm$ 0.36 & 83.29 $\pm$ 0.43 & 83.11 $\pm$ 0.48 & +3.37\\
    & BL-IST & BIH & 69/141 & 99.95 $\pm$ 0.03 & 99.85 $\pm$ 0.02 & 96.41 $\pm$ 0.11 & 96.11 $\pm$ 0.16 & +1.79 & 261/775 & 99.98 $\pm$ 0.02 & 98.69 $\pm$ 0.26 & 82.86 $\pm$ 0.50 & 83.03 $\pm$ 0.43 & +3.29\\
    & BL-IST & VBI$_\text{iter1}$ & 79/231 & 99.99 $\pm$ 0.01 & 99.89 $\pm$ 0.02 & 96.61 $\pm$ 0.10 & 96.18 $\pm$ 0.26 & +1.86 & 304/775 & 99.99 & 99.02 $\pm$ 0.21 & 83.59 $\pm$ 0.46 & 83.00 $\pm$ 0.17 & +3.26\\
    & BL-IST & VBI & 130/131 & 99.99 $\pm$ 0.01 & 99.96 $\pm$ 0.01 & 96.56 $\pm$ 0.12 & 96.50 $\pm$ 0.18 & +2.18 & 741/775 & 99.98 $\pm$ 0.02 & 99.57 $\pm$ 0.14 & 82.96 $\pm$ 0.23 & 82.87 $\pm$ 0.07 & +3.13\\
    & BL-IST & DDN & 131/131 & 99.97 $\pm$ 0.04 & 99.97 $\pm$ 0.04 & 96.61 $\pm$ 0.27 & 96.35 $\pm$ 0.12 & +2.03 & 775/775 & 99.98 $\pm$ 0.01 & 99.98 $\pm$ 0.01 & 83.29 $\pm$ 0.42 & 83.03 $\pm$ 0.07 & +3.29 \\
    & BL-IST & SP & 17/131 & 99.97 $\pm$ 0.01 & 99.82 $\pm$ 0.01 & 96.61 $\pm$ 0.22 & 96.30 $\pm$ 0.15 & +1.98 & 45/775 & 99.99 $\pm$ 0.01 & 98.58 $\pm$ 0.35 & 83.00 $\pm$ 0.19 & 83.25 $\pm$ 0.32 & +3.51\\
    \midrule
    \multirow{8}{*}{\makecell{EfficientNetV2-M \\(52.99M)}} & BL & - & - & - & 99.96 $\pm$ 0.06 & 97.66 $\pm$ 0.13 & 97.15 $\pm$ 0.14 & - & - & - & 99.88 $\pm$ 0.08 & 86.63 $\pm$ 0.73 & 86.88 $\pm$ 0.46 & -\\
    & BL-IST & None & - & 99.97 $\pm$ 0.01 & 99.95 $\pm$ 0.01 & 98.21 $\pm$ 0.08 & 97.76 $\pm$ 0.14 & +0.61 & - & 99.72 $\pm$ 0.11 & 99.62 $\pm$ 0.14 & 87.73 $\pm$ 0.59 & 87.36 $\pm$ 0.57 & +0.48 \\
    & BL-IST & LL & 3/9 & 100.00 & 99.98 & 98.14 $\pm$ 0.09 & 97.82 $\pm$ 0.08 & +0.67 & 17/54 & 99.95 $\pm$ 0.05 & 99.87 $\pm$ 0.01 & 88.05 $\pm$ 0.22 & 87.52 $\pm$ 0.45 & +0.64\\
    & BL-IST & BIH & 6/9 & 100.00 & 99.99 $\pm$ 0.01 & 98.20 $\pm$ 0.09 & 97.82 $\pm$ 0.09 & +0.68 & 23/54 & 99.97 $\pm$ 0.05 & 99.91 $\pm$ 0.09 & 87.96 $\pm$ 0.32 & \textbf{88.00 $\pm$ 0.10} & \textbf{+1.12}\\
    & BL-IST & VBI$_\text{iter1}$ & 7/9 & 99.99 $\pm$ 0.01 & 99.99 $\pm$ 0.01 & 98.18 $\pm$ 0.11 & 97.82 $\pm$ 0.12 & +0.67 & 29/54 & 99.94 & 99.87 $\pm$ 0.05 & 88.00 $\pm$ 0.03 & 87.77 $\pm$ 0.06 & +0.89\\
    & BL-IST & VBI & 8/9 & 99.99 $\pm$ 0.01 & 99.99 $\pm$ 0.01 & 98.13 $\pm$ 0.12 & 97.80 $\pm$ 0.04 & +0.65 & 46/54 & 99.95 $\pm$ 0.01 & 99.88 $\pm$ 0.04 & 88.09 $\pm$ 0.18 & 87.76 $\pm$ 0.16 & +0.88\\
    & BL-IST & DDN & 9/9 & 100.00 & 100.00 & 98.18 $\pm$ 0.09 & \textbf{97.86 $\pm$ 0.06} & \textbf{+0.71} & 54/54 & 99.92 $\pm$ 0.04 & 99.92 $\pm$ 0.04 & 87.98 $\pm$ 0.18 & 87.81 $\pm$ 0.10 & +0.93 \\
    & BL-IST & SP & 4/9 & 99.99 & 99.98 & 98.13 $\pm$ 0.05 & 97.70 $\pm$ 0.12 & +0.55 &18/54 & 99.95 $\pm$ 0.03 & 99.87 $\pm$ 0.07 & 87.85 $\pm$ 0.04 & 87.89 $\pm$ 0.19 & +1.01\\\bottomrule
\end{tabular}
}
\caption{Accuracy (\%) of FP32 baseline models (BL), which is fine-tuned on the CIFAR train domains, and accuracy of baselines after applying our approach (denoted as BL-IST) by using different attacks to generate adversarial domains. The data is reported as an average of three seeds.} 
\label{table:resnet_fp_cifar_3seed}
\end{table*}

\subsection{Adversarial Correction of FP32 Models}
\begin{figure}[hbt!] 
         \centering
     \begin{subfigure}[b]{0.23\textwidth}
         \centering
        \includegraphics[width=\textwidth]{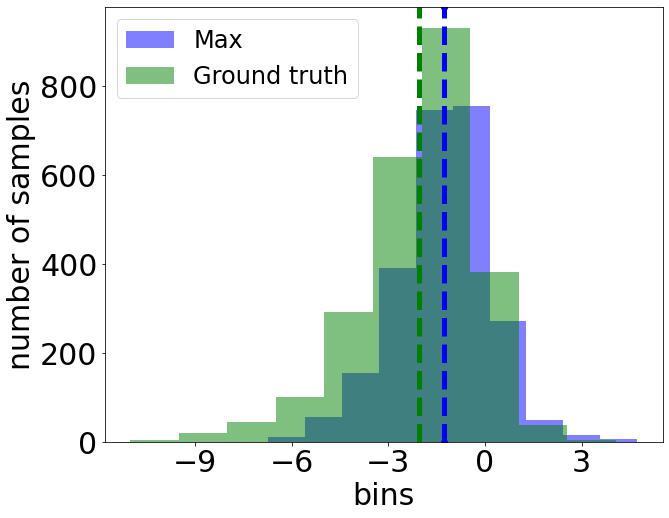}
        \caption{Incorrect samples, LL}
    \label{incorr_img_softmax0_resnet34_cifar100_ILLC_prob_dist_diff}
     \end{subfigure}
     \begin{subfigure}[b]{0.23\textwidth}
         \centering
        \includegraphics[width=\textwidth]{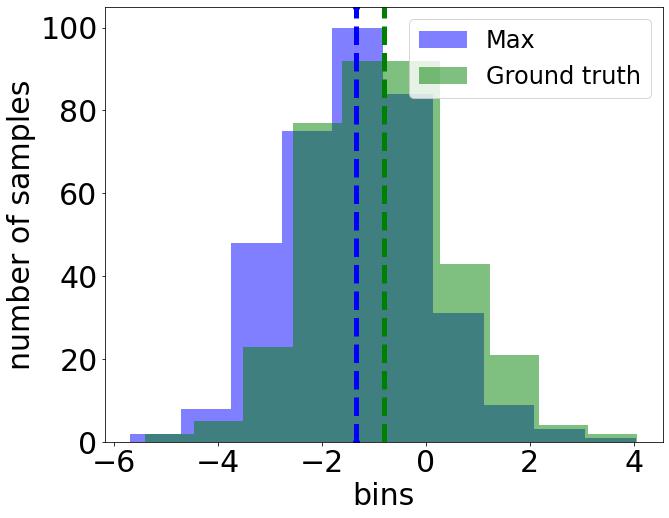}
        \caption{Corrected samples, LL}
        \label{corrected_img_softmax0_resnet34_cifar100_ILLC_prob_dist_diff}
     \end{subfigure} \\
    \begin{subfigure}[b]{0.23\textwidth}
         \centering
        \includegraphics[width=\textwidth]{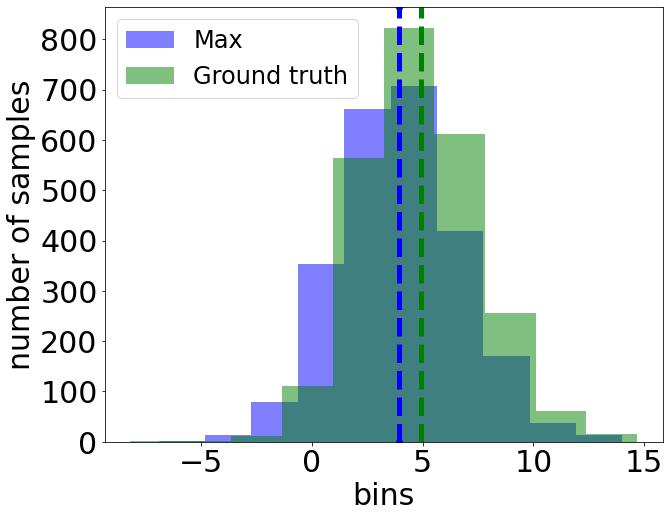}
        \caption{Incorrect samples, VBI}
    \label{incorr_img_softmax0_resnet34_cifar100_VBI_prob_dist_diff}
     \end{subfigure}
     \begin{subfigure}[b]{0.23\textwidth}
         \centering
        \includegraphics[width=\textwidth]{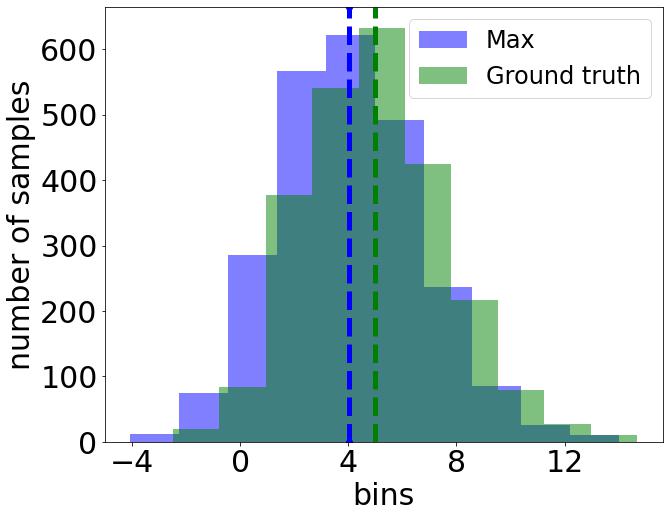}
        \caption{Corrected samples, VBI}
    \label{corrected_img_softmax0_resnet34_cifar100_VBI_prob_dist_diff}
     \end{subfigure} \\
     \caption{The incorrect class (max) and true class logits change for uncorrected (a,c) and corrected (b,d) samples of CIFAR-100 after applying the corrective LL (a,b) and VBI (c,d) attacks on the ResNet-34. The vertical dashed lines indicate mean values of incorrect class (max logit) and true class logits change.}
    \label{fig:plt_pb_hist_resnet34_cifar100_LL}
\end{figure}

The training, validation, and test accuracies of various networks obtained by applying AdCorDA for different attack methods on CIFAR-10 and CIFAR-100 are shown in \cref{table:resnet_fp_cifar_3seed}. The ``None" attack case corresponds to the situation where we do not apply any adversarial correction, effectively relying only on the curriculum modification of the training set. Our approach enhances the model performance by as much as 2.68\% and 5.23\% on CIFAR-10 and CIFAR-100, respectively, when utilizing ResNets of various sizes. As for the effect of our method when applied to EfficientNet, we observe an improvement of about 0.7-1.1\% on CIFAR datasets. More specifically, the ResNet-34 baseline model, operating at full precision, achieved a test accuracy of 78.41\% on CIFAR-100. Our adversarial correction method, using a DDN adversarial attack, improves the test accuracy to 83.64\%, representing a notable increase of 5.23\%. 
Upon incorporating adversarial correction using the LL adversarial attack on the training set, we observed a decrease in the initial training loss from 0.254 (on the original training set $T$) to 0.173 (on the corrected training set $T^\prime$) on CIFAR-100.
This shows that the adversarial correction does indeed reduce the training loss. 
In \cref{corrected_img_softmax0_resnet34_cifar100_ILLC_prob_dist_diff} and \cref{corrected_img_softmax0_resnet34_cifar100_VBI_prob_dist_diff} we can see that both targeted (VBI) and untargeted (LL) adversarial attacks can successfully reduce the logit level of the initially maximum probability incorrect label as compared with the logit level of the true label, resulting in correction. 

\subsection{Adversarial Correction of Quantized Models}
Table \ref{table:resnet_qt_cifar} shows that our method also improves the baseline performance of quantized networks. For example, the full precision baseline ResNet-34 achieves a test accuracy of 78.41\% on CIFAR-100. The Int8 quantized baseline ResNet34 has a test accuracy of 77.13\% on CIFAR-100. When applying our method using the BIH adversarial attack on the full precision baseline, we achieve a test accuracy of 82.99\%, representing an improvement of +4.58\%. After Int8 PTSQ quantization, the modified FP network achieves a test accuracy of 82.18\% - an improvement of +5.05\% over the original quantized network (and an improvement of +3.77\% over the original full precision network!).

\begin{table*}[hbt!]
\centering
\resizebox{\textwidth}{!}{%
\begin{tabular}{cccccccccccccccccc} \toprule
    \multirow{2.5}{*}{Model} & \multirow{2.5}{*}{Approach} & \multirow{2.5}{*}{Attack} & \multicolumn{6}{c}{CIFAR-10} & \multicolumn{6}{c}{CIFAR-100} \\ \cmidrule(lr){4-9} \cmidrule(lr){10-15} 
    & & & Corr. rate & $T^\prime$ & Train & Valid & Test & $\Delta$ Acc & Corr. rate & $T^\prime$ & Train & Valid & Test & $\Delta$ Acc\\\midrule
    \multirow{8}{*}{ResNet-18} & BL & - & - & - & 99.61 $\pm$ 0.56 & 93.73 $\pm$ 0.43 & 93.29 $\pm$ 0.37 & - & - & - & 99.00 $\pm$ 1.19 & 76.84 $\pm$ 0.12 & 77.04 $\pm$ 0.08 & -\\
    & PTSQ & - & - & - & 98.08 $\pm$ 0.71 & 93.01 $\pm$ 0.56 & 92.42 $\pm$ 0.17 & - & - & - & 96.74 $\pm$ 3.02 & 75.45 $\pm$ 1.29 & 76.06 $\pm$ 0.94 & -\\
    & PTSQ-IST (bef. qt) & None & - & 99.96 $\pm$ 0.04 & 99.43 $\pm$ 0.02 & 95.88 $\pm$ 0.26 & 95.59 $\pm$ 0.07 & - & - & 99.80 $\pm$ 0.12 & 97.13 $\pm$ 0.57 & 80.34 $\pm$ 0.59 & 80.18 $\pm$ 0.39 & -\\
    & PTSQ-IST (aft. qt) & None & - & 99.93 $\pm$ 0.01 & 99.23 $\pm$ 0.05 & 95.30 $\pm$ 0.39 & 95.18 $\pm$ 0.09 & +2.76 & - & 99.52 $\pm$ 0.14 & 96.67 $\pm$ 0.56 & 79.15 $\pm$ 0.53 & 79.15 $\pm$ 0.26 & +3.09\\
    & PTSQ-IST (bef. qt) & BIH & 158/736 & 100.00 & 99.53 & 96.07 $\pm$ 0.06 & 95.85 $\pm$ 0.19 & -  & 300/1966 & 100.00 & 97.55 $\pm$ 0.50 & 80.61 $\pm$ 0.16 & 80.82 $\pm$ 0.30 & -\\
    & PTSQ-IST (aft. qt) & BIH & - & 99.99 $\pm$ 0.01 & 99.46 $\pm$ 0.04 & 95.53 $\pm$ 0.29 & \textbf{95.48 $\pm$ 0.18} & \textbf{+3.07} & - & 99.98 $\pm$ 0.02 & 97.36 $\pm$ 0.55 & 79.25 $\pm$ 0.30 & 79.53 $\pm$ 0.58 & +3.47 \\
    & PTSQ-IST (bef. qt) & SP & 128/736 & 100.00 & 99.54 $\pm$ 0.03 & 96.17 $\pm$ 0.13 & 95.72 $\pm$ 0.22 & - & 189/1966 & 100.00 & 97.17 $\pm$ 0.62 & 80.40 $\pm$ 0.45 & 81.07 $\pm$ 0.23 & -\\
    & PTSQ-IST (aft. qt) & SP & - & 100.00 & 99.48 $\pm$ 0.02 & 95.46 $\pm$ 0.20 & 95.29 $\pm$ 0.06 & +2.93 & - & 99.98 $\pm$ 0.01 & 97.31 $\pm$ 0.66 & 79.27 $\pm$ 0.73 & \textbf{79.79 $\pm$ 0.49} & \textbf{+3.73}\\\midrule
    \multirow{8}{*}{ResNet-34} & BL & - & - & - & 99.43 $\pm$ 0.67 & 94.71 $\pm$ 0.05 & 94.22 $\pm$ 0.06 & - & - & - & 94.36 $\pm$ 2.24 & 78.12 $\pm$ 0.79 & 78.41 $\pm$ 0.10 & - \\
    & PTSQ & - & - & - & 98.16 $\pm$ 0.35 & 93.63 $\pm$ 0.10 & 93.36 $\pm$ 0.09 & - & - & - & 90.32 $\pm$ 2.30 & 76.20 $\pm$ 0.39 & 77.13 $\pm$ 0.45 & -\\
    & PTSQ-IST (bef. qt) & None & - & 99.97 $\pm$ 0.02 & 99.44 $\pm$ 0.15 & 96.57 $\pm$ 0.15 & 96.28 $\pm$ 0.13 & - & - & 99.09 $\pm$ 0.10 & 93.03 $\pm$ 0.29 & 83.08 $\pm$ 0.23 & 82.94 $\pm$ 0.29 & -\\
    & PTSQ-IST (aft. qt) & None & - & 99.92 $\pm$ 0.04 & 99.31 $\pm$ 0.17 & 96.19 $\pm$ 0.10 & \textbf{96.08 $\pm$ 0.20} & \textbf{+2.72} & - & 99.44 $\pm$ 0.55 & 92.59 $\pm$ 0.33 & 81.89 $\pm$ 0.63 & 81.94 $\pm$ 0.45 & +4.81\\
    & PTSQ-IST (bef. qt) & BIH & 250/771 & 100.00 & 99.45 $\pm$ 0.11 & 96.61 $\pm$ 0.19 & 96.33 $\pm$ 0.15 & - & 689/4607 & 99.96 $\pm$ 0.04 & 93.97 $\pm$ 0.31 & 83.20 $\pm$ 0.24 & 82.99 $\pm$ 0.15 & - \\
    & PTSQ-IST (aft. qt) & BIH & - & 99.97 $\pm$ 0.02 & 99.39 $\pm$ 0.12 & 96.29 $\pm$ 0.08 & 96.05 $\pm$ 0.07 & +2.69 & - & 99.99 $\pm$ 0.01 & 93.77 $\pm$ 0.33 & 82.19 $\pm$ 0.14 & \textbf{82.18 $\pm$ 0.20} & \textbf{+5.05}\\
    & PTSQ-IST (bef. qt) & SP & 272/771 & 99.96 $\pm$ 0.04 & 99.51 $\pm$ 0.04 & 96.58 $\pm$ 0.14 & 96.12 $\pm$ 0.23 & - &  479/4607 & 100.00 & 93.80 $\pm$ 0.52 & 83.13 $\pm$ 0.54 & 82.90 $\pm$ 0.19 & -  \\
    & PTSQ-IST (aft. qt) & SP & - & 99.95 $\pm$ 0.05 & 99.45 $\pm$ 0.05 & 96.25 $\pm$ 0.06 & 95.83 $\pm$ 0.19 & +2.47 & - & 100.00 & 93.55 $\pm$ 0.53 & 81.99 $\pm$ 0.24 & 82.12 $\pm$ 0.20 & +4.99\\
    \bottomrule
\end{tabular}}
\caption{Accuracy (\%) of quantized (Int8) ResNets of various sizes obtained after applying PTSQ on its baseline, and the accuracy of Int8 ResNets using our approach.}
\label{table:resnet_qt_cifar}
\end{table*}

It is also worth noting in \cref{table:resnet_qt_cifar}, that the decrease in accuracy on ResNet-34 after quantization using our method is only 0.81\% (i.e., from 82.99\% to 82.18\%), whereas the drop in performance in the original network after quantization is 1.28\% (i.e., from 78.41\% to 77.13\%). This comparison shows that our adversarial correction method makes the full-precision network less affected by quantization.

The quantized ResNet-34 network after using our adversarial correction technique achieves a higher accuracy (82.18\%) than even that of a normally trained full-precision ResNet-152 baseline model (81.52\%), while significantly reducing the model size (20.76MB vs 223.49MB).

\subsection{Adversarial Perturbation vs. Correction}

In the Feng and Tu theory, all that is needed in the first step of the IST is to perturb the input so as to reduce the loss. It is not necessary to actually change the input so as to have the network give the correct answer; all that is required is that the loss be reduced.

In the experiments shown in \cref{table:resnet_fp_cifar_3seed}, we defined $T_a$ as the set of successfully corrected samples in step 3 of our adversarial correction approach. If we now consider $T_a$ to include all perturbed samples, whether the outputs are corrected or not, $T'$ will have the same size as the original training dataset. We refer to the network adapted using this variation as \textit{BL-IST-A}. In our original approach, the accuracy of the original network on $T'$ reaches 100\% when we consider only the successfully perturbed samples and the original correctly detected samples. Inspired by \cite{Shen2023Fast}, we can think of $T'$ as an \emph{easy} dataset, given its 100\% accuracy, while considering $T$ as a \emph{hard} dataset. In Table \ref{table:resnet_fp_cifar_mod} we observe a drop in performance improvement in BL-IST-A as compared to our first approach. This could be attributed to the adversarial perturbations increasing the loss rather than decreasing it, as compared with the baseline, for the uncorrected inputs. We conclude that we should only retain the corrected input samples.


\begin{table}[hbt!]
\centering
\resizebox{\columnwidth}{!}{%
\begin{tabular}{cccccccc} \toprule
\multirow{2.5}{*}{Model} & \multirow{2.5}{*}{Approach} & \multicolumn{3}{c}{CIFAR-10} & \multicolumn{3}{c}{CIFAR-100} \\ \cmidrule(lr){3-5} \cmidrule(lr){6-8} 
    & & \# $T^\prime$ & Test & $\Delta$ Acc & \# $T^\prime$ & Test & $\Delta$ Acc \\
\midrule
\multirow{3}{*}{\makecell{ResNet-18}} 
    & BL       & - & 93.32 & - & - & 77.09 & - \\
    & BL-IST   & 44,972 & 95.77 & +2.45 & 44,879 & 80.48 & +3.39\\
    & BL-IST-A & 45,000 & 95.51 & +2.19 & 45,000 & 79.56 & +2.47\\
\midrule
\multirow{3}{*}{\makecell{ResNet-34}} 
    & BL       & - & 94.24 & - & - & 78.53 & - \\
    & BL-IST   & 44,993 & 96.36 & +2.12 & 42,903 & 82.76 & +4.23\\
    & BL-IST-A & 45,000 & 96.18 & +1.94 & 45,000 & 80.81 & +2.28 \\
\bottomrule
\end{tabular}
}
\caption{Accuracy (\%) of ResNet FP32 baselines after applying our approach using the LL attack to generate adversarial domains for CIFAR datasets. Note that BL-IST-A is a refined approach in which $T_a$ in Step 3 incorporates all perturbed samples of $T_w$.}
\label{table:resnet_fp_cifar_mod}
\end{table}

%

\subsection{Grad-CAM visualization of Adversarial Correction}
\begin{figure}[hbt!]
    \centering
    \begin{subfigure}[b]{0.19\textwidth}
         \centering
        \includegraphics[width=\textwidth]{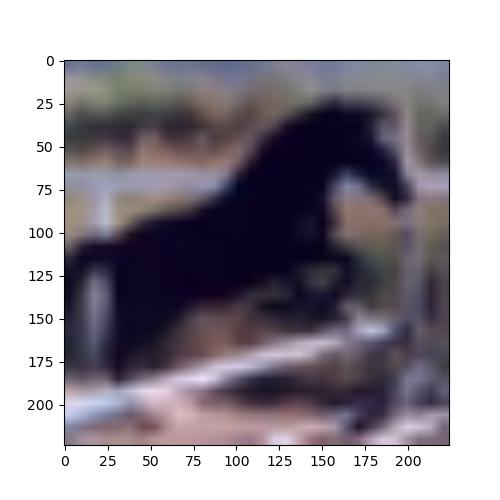}
        \caption{}
        \label{img:incorr6}
     \end{subfigure}
     \begin{subfigure}[b]{0.28\textwidth}
         \centering
        \includegraphics[width=\textwidth]{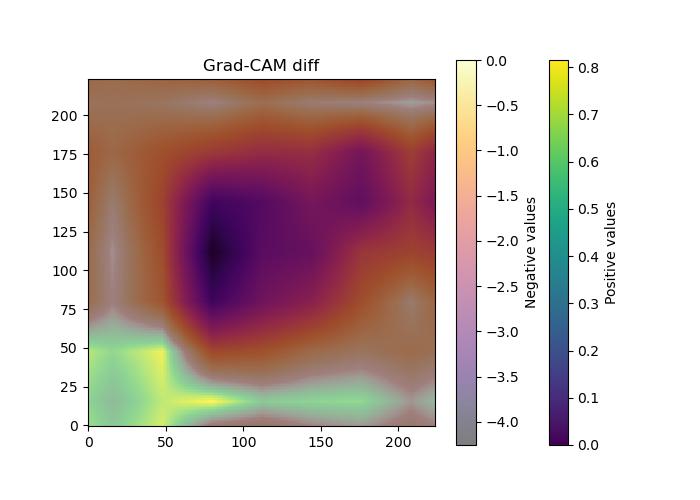}
        \caption{}
        \label{img:incorr6_gradcam_diff}
     \end{subfigure}
     \begin{subfigure}[b]{0.23\textwidth}
         \centering
         \includegraphics[width=\textwidth]{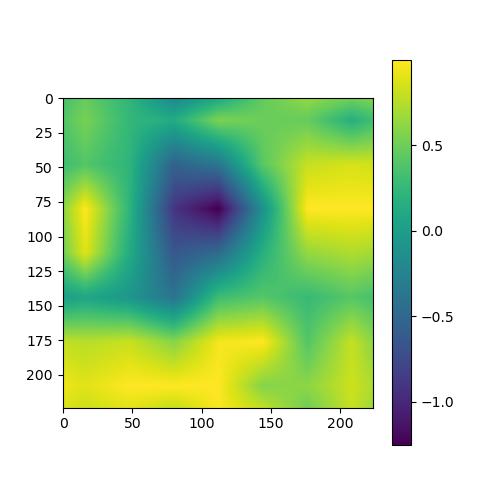}
         \caption{}
         \label{img:incorr6_gradcam}
     \end{subfigure}
     \begin{subfigure}[b]{0.23\textwidth}
         \centering
         \includegraphics[width=\textwidth]{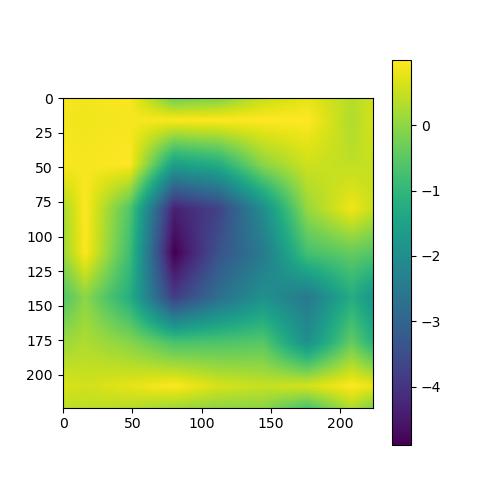}
         \caption{}
         \label{img:corr6_gradcam}
     \end{subfigure}
     \caption{Evaluation of ResNet-34 on CIFAR-10 dataset. (a) misclassified images, (b) the difference between the Grad-CAM images for the original and adversarially corrected inputs using DDN attack. This illustrates the shift in focus of the network for the two images, (c) the Grad-CAM image for the original incorrect image, (d) the Grad-CAM image for the adversarially corrected image.}
    \label{fig:plt_gradcam}
\end{figure}

To help visualize the impact of the adversarial correction technique on misclassified images, we employ Gradient-weighted Class Activation Mapping (Grad-CAM) \cite{Selvaraju2017GradCAM} to provide visual explanations. Grad-CAM utilizes gradient-based localization to identify important regions in an image that contribute to the model's concept prediction. In our study, \cref{img:incorr6} is an example initially misclassified as an `automobile' by ResNet-34. However, applying the DDN attack, the image can be correctly identified as a `horse'. To better understand the differences between the Grad-CAM of the original (\cref{img:incorr6_gradcam}) and its corrected image (\cref{img:corr6_gradcam}), we present a visualization in \cref{img:incorr6_gradcam_diff}. This visualization clearly illustrates that the incorrect detection was primarily influenced by the surrounding contextual information rather than the object itself. This demonstrates that by modifying the surrounding contextual information of the image using the adversarial attack, correct classification becomes possible. 

\begin{table}[hbt!]
\centering
\resizebox{\columnwidth}{!}{%
\begin{tabular}{ccccccc} \toprule
\multirow{2.5}{*}{Model} & \multirow{2.5}{*}{Approach} & \multirow{2.5}{*}{\makecell{Attack}} & \multicolumn{2}{c}{CIFAR-10} & \multicolumn{2}{c}{CIFAR-100} \\ \cmidrule(lr){4-5} \cmidrule(lr){6-7} 
    & & & Clean & AutoAttack & Clean & AutoAttack \\
\midrule
\multirow{4}{*}{\makecell{ResNet-18}} 
    & BL      &   -  &  $93.29_{\pm 0.37}$ & $15.92_{\pm 1.67}$ & $77.04_{\pm 0.08}$ & $7.56_{\pm 1.14}$ \\
    & BL-IST & None & $95.57_{\pm 0.13}$ & $47.63_{\pm 1.74}$ & $80.27_{\pm 0.74}$ & $20.65_{\pm 0.82}$ \\
    & BL-IST & DDN  & $\mathbf{95.84}_{\pm 0.07}$ & $47.97_{\pm 0.10}$ & $80.82_{\pm 0.35 }$ & $21.66_{\pm 0.94}$ \\
    & BL-IST & SP   & $95.80_{\pm 0.08}$ & $\mathbf{50.97}_{\pm 0.72}$ & $\mathbf{80.89}_{\pm 0.61}$ & $\mathbf{21.80}_{\pm 1.87}$ \\
\midrule
\multirow{4}{*}{\makecell{ResNet-34}} 
    & BL      &  -   & $94.22_{\pm 0.06}$ & $13.80_{\pm 1.05}$ & $78.41_{\pm 0.10}$ & $7.90_{\pm 0.62}$ \\
    & BL-IST & None & $96.40_{\pm 0.05}$ & $50.54_{\pm 2.90}$ & $82.98_{\pm 0.07}$ & $22.37_{\pm 1.39}$ \\
    & BL-IST & DDN  & $\mathbf{96.71}_{\pm 0.05}$ & $\mathbf{51.03}_{\pm 2.89}$ & $\mathbf{83.64}_{\pm 0.06}$ & $20.68_{\pm 2.19}$ \\
    & BL-IST & SP   & $96.22_{\pm 0.05}$ & $50.13_{\pm 2.41}$ & $83.25_{\pm 0.29}$ & $\mathbf{24.47}_{\pm 0.31}$ \\
\midrule
\multirow{4}{*}{\makecell{EfficientNetV2-M}} 
    & BL      &  -   & $97.15_{\pm 0.14}$ & $15.07_{\pm 0.78}$ & $86.88_{\pm 0.46}$ & $11.16_{\pm 0.45}$ \\
    & BL-IST & None & $97.76_{\pm 0.14}$ & $\mathbf{52.68}_{\pm 3.20}$ & $87.36_{\pm 0.57}$ & $23.61_{\pm 3.08}$\\
    & BL-IST & DDN  & $\mathbf{97.86}_{\pm 0.06}$ & $42.42_{\pm 2.66}$ & $87.81_{\pm 0.10}$ & $25.72_{\pm 2.15}$ \\
    & BL-IST & SP   & $97.70_{\pm 0.12}$ & $39.02_{\pm 2.36}$ & $\mathbf{87.89}_{\pm 0.19}$ & $\mathbf{25.84}_{\pm 1.79}$ \\
\bottomrule
\end{tabular}
}
\caption{Accuracy (\%) of FP32 baselines and adapted models using our approach on the clean and adversarially perturbed CIFAR test sets. AutoAttack is used to generate the adversarial samples.}
\label{table:Adv_robustness}
\end{table}

\subsection{Enhanced Robustness to Adversarial Attacks}
Our adversarial correction technique has many similarities to \emph{adversarial training} methods for enhancing robustness to adversarial attacks. Such methods generate adversarial examples, for which networks give the wrong answer, and add these as augmentations of the original dataset. Fine-tuning on the augmented dataset leads to enhanced robustness against adversarial attacks \cite{madry2017towards}. Our approach is similar in that we create new images resulting from adversarial attacks, and use these in concert with images from the original dataset in further training. There are significant differences, however, between our method and standard adversarial training. First, we do not augment the original dataset, but instead replace some of the samples in the original dataset with the adversarial examples. Second, the adversarial attacks are only applied to samples that the network gets wrong, rather than samples that the network gets right, and we only keep the adversarial examples which are corrective - that the network now gets right. Finally, rather than doing fine-tuning using standard training on the augmented training set, we do domain adaptation from the adversarially corrected training set to the original training set. We tested the robustness of ResNets to the AutoAttack suite of attacks
\cite{AutoAttack}. As seen in \cref{table:Adv_robustness}, our method provides significant robustness to adversarial attacks. For CIFAR-10 with ResNet-18 we see an improvement from 15.92\% on the baseline model to 50.97\% on the adversarially corrected model with the SP correction method. On CIFAR-100 with ResNet-18 we see an improvement from 7.56\% to 21.8\%. Note that using only curriculum domain adaptation (the ``None" case) also gives significant robustness. While current state-of-the-art robust network techniques get higher accuracies under attack than ours (e.g., 27.67\% on CIFAR-100 by \cite{addepalli2022efficient} and 55.54\% on CIFAR-10 by \cite{sehwag2021robust}, both with ResNet-18), our focus is on attaining higher clean (before attacks) accuracies, and the enhanced robustness is a welcome byproduct. Jointly optimizing both clean accuracy and adversarial robustness is an interesting avenue for future work.
\section{Conclusion}
\label{sec:conclusion}

In this work, we present a new method for enhancing the performance of trained image classifier networks. The method has two stages - first the training set samples for which the network gives incorrect answers are modified via corrective adversarial attacks so that the network now gives the correct answers. In the second stage, the network is refined via domain adaptation, using Deep CORAL, from the modified dataset to the original dataset. Experiments show substantial enhancements in performance on CIFAR-10 and CIFAR-100, of over 4\%. 

One could argue that in doing adversarial correction we are performing a type of dataset augmentation, by creating new samples with known labels. However, we are not training on this augmented dataset in a standard manner. Instead, the removal of the incorrect samples and the addition of the corrected samples provides a more pure representation of the domain that the initial network does well on, thereby enhancing the effectiveness of the subsequent domain adaptation step. Indeed, even just removing the incorrect samples, without adding the adversarial corrections, provides a significant benefit to the domain adaptation step.

Our experiments show that the adversarial correction approach is effective for refining quantized networks. Also, we observe that the adversarial correction enhances robustness to adversarial attack.
{
    \small
    \bibliographystyle{ieeenat_fullname}
    \bibliography{main}

\begin{thebibliography}{30}
\providecommand{\natexlab}[1]{#1}
\providecommand{\url}[1]{\texttt{#1}}
\expandafter\ifx\csname urlstyle\endcsname\relax
  \providecommand{\doi}[1]{doi: #1}\else
  \providecommand{\doi}{doi: \begingroup \urlstyle{rm}\Url}\fi

\bibitem[Addepalli et~al.(2022)Addepalli, Jain, et~al.]{addepalli2022efficient}
Sravanti Addepalli, Samyak Jain, et~al.
\newblock Efficient and effective augmentation strategy for adversarial training.
\newblock \emph{Advances in Neural Information Processing Systems}, 35:\penalty0 1488--1501, 2022.

\bibitem[Andriushchenko et~al.(2020)Andriushchenko, Croce, Flammarion, and Hein]{SquareAttack}
Maksym Andriushchenko, Francesco Croce, Nicolas Flammarion, and Matthias Hein.
\newblock Square attack: {A} query-efficient black-box adversarial attack via random search.
\newblock In \emph{Computer Vision - {ECCV} 2020 - 16th European Conference, Glasgow, UK, August 23-28, 2020, Proceedings, Part {XXIII}}, pages 484--501. Springer, 2020.

\bibitem[Bengio et~al.(2009)Bengio, Louradour, Collobert, and Weston]{Bengio_curriculumlearning}
Yoshua Bengio, Jérôme Louradour, Ronan Collobert, and Jason Weston.
\newblock Curriculum learning.
\newblock In \emph{International Conference on Machine Learning}, pages 41--48, 2009.

\bibitem[Bottou(2010)]{SGD}
Léon Bottou.
\newblock Large-scale machine learning with stochastic gradient descent.
\newblock In \emph{Proceedings in Computational Statistics}, pages 177--186. Physica-Verlag HD, 2010.

\bibitem[Croce and Hein(2020{\natexlab{a}})]{AutoAttack}
Francesco Croce and Matthias Hein.
\newblock Reliable evaluation of adversarial robustness with an ensemble of diverse parameter-free attacks.
\newblock In \emph{International Conference on Machine Learning}, pages 2206--2216. PMLR, 2020{\natexlab{a}}.

\bibitem[Croce and Hein(2020{\natexlab{b}})]{FAB}
Francesco Croce and Matthias Hein.
\newblock Minimally distorted adversarial examples with a fast adaptive boundary attack.
\newblock In \emph{International Conference on Machine Learning}, pages 2196--2205. PMLR, 2020{\natexlab{b}}.

\bibitem[Deng et~al.(2009)Deng, Dong, Socher, Li, Li, and Fei-Fei]{ImageNet}
Jia Deng, Wei Dong, Richard Socher, Li-Jia Li, Kai Li, and Li Fei-Fei.
\newblock {ImageNet}: {A} large-scale hierarchical image database.
\newblock In \emph{IEEE Conference on Computer Vision and Pattern Recognition}, pages 248--255, 2009.

\bibitem[Feng and Tu(2022)]{feng2022activityweight}
Yu Feng and Yuhai Tu.
\newblock The activity-weight duality in feed forward neural networks: {T}he geometric determinants of generalization.
\newblock \emph{arXiv preprint arXiv:2203.10736}, 2022.

\bibitem[Goodfellow et~al.(2015)Goodfellow, Shlens, and Szegedy]{Goodfellow2015Explaining}
Ian~J. Goodfellow, Jonathon Shlens, and Christian Szegedy.
\newblock Explaining and harnessing adversarial examples.
\newblock In \emph{International Conference on Learning Representations}, 2015.

\bibitem[He et~al.(2016)He, Zhang, Ren, and Sun]{he2015deep}
Kaiming He, Xiangyu Zhang, Shaoqing Ren, and Jian Sun.
\newblock Deep residual learning for image recognition.
\newblock In \emph{IEEE Conference on Computer Vision and Pattern Recognition}, pages 770--778, 2016.

\bibitem[Jacob et~al.(2018)Jacob, Kligys, Chen, Zhu, Tang, Howard, et~al.]{Jacob2018Quantization}
Benoit Jacob, Skirmantas Kligys, Bo Chen, Menglong Zhu, Matthew Tang, Andrew Howard, et~al.
\newblock Quantization and training of neural networks for efficient integer-arithmetic-only inference.
\newblock In \emph{IEEE/CVF Conference on Computer Vision and Pattern Recognition}, pages 2704--2713, 2018.

\bibitem[Krizhevsky and Hinton(2009)]{Krizhevsky2009Learning}
Alex Krizhevsky and Geoffrey Hinton.
\newblock Learning multiple layers of features from tiny images.
\newblock Technical report, University of Toronto, 2009.

\bibitem[Kumar et~al.(2010)Kumar, Packer, and Koller]{kumar2010self}
M Kumar, Benjamin Packer, and Daphne Koller.
\newblock Self-paced learning for latent variable models.
\newblock \emph{Advances in neural information processing systems}, 23, 2010.

\bibitem[Kurakin et~al.(2017)Kurakin, Goodfellow, and Bengio]{Kurakin2017Adversarial}
Alexey Kurakin, Ian~J. Goodfellow, and Samy Bengio.
\newblock Adversarial examples in the physical world.
\newblock In \emph{International Conference on Learning Representations}. OpenReview.net, 2017.

\bibitem[Li et~al.(2022)Li, Cheng, Hsieh, and Lee]{Li2022ReviewAdversarial}
Yao Li, Minhao Cheng, Cho-Jui Hsieh, and Thomas C.~M. Lee.
\newblock A review of adversarial attack and defense for classification methods.
\newblock \emph{The American Statistician}, 76\penalty0 (4):\penalty0 329--345, 2022.

\bibitem[Madry et~al.(2017)Madry, Makelov, Schmidt, Tsipras, and Vladu]{madry2017towards}
Aleksander Madry, Aleksandar Makelov, Ludwig Schmidt, Dimitris Tsipras, and Adrian Vladu.
\newblock Towards deep learning models resistant to adversarial attacks.
\newblock \emph{arXiv preprint arXiv:1706.06083}, 2017.

\bibitem[Madry et~al.(2018)Madry, Makelov, Schmidt, Tsipras, and Vladu]{Mary2018Towards}
Aleksander Madry, Aleksandar Makelov, Ludwig Schmidt, Dimitris Tsipras, and Adrian Vladu.
\newblock Towards deep learning models resistant to adversarial attacks.
\newblock In \emph{International Conference on Learning Representations}, 2018.

\bibitem[Paszke et~al.(2019)Paszke, Gross, Massa, Lerer, Bradbury, Chanan, et~al.]{paszke_pytorch_2019}
Adam Paszke, Sam Gross, Francisco Massa, Adam Lerer, James Bradbury, Gregory Chanan, et~al.
\newblock {PyTorch}: {An} imperative style, high-performance deep learning library.
\newblock In \emph{Advances in {Neural} {Information} {Processing} {Systems}}. Curran Associates, Inc., 2019.

\bibitem[Rauber et~al.(2017)Rauber, Brendel, and Bethge]{rauber2017foolbox}
Jonas Rauber, Wieland Brendel, and Matthias Bethge.
\newblock Foolbox: {A} python toolbox to benchmark the robustness of machine learning models.
\newblock In \emph{International Conference on Machine Learning Workshop}, 2017.

\bibitem[Rauber et~al.(2020)Rauber, Zimmermann, Bethge, and Brendel]{rauber2017foolboxnative}
Jonas Rauber, Roland Zimmermann, Matthias Bethge, and Wieland Brendel.
\newblock Foolbox native: {F}ast adversarial attacks to benchmark the robustness of machine learning models in {PyTorch}, {TensorFlow}, and {JAX}.
\newblock \emph{Journal of Open Source Software}, 5\penalty0 (53):\penalty0 2607, 2020.

\bibitem[Rony et~al.(2019)Rony, Hafemann, Oliveira, Ayed, Sabourin, and Granger]{rony2019decoupling}
J{\'e}r{\^o}me Rony, Luiz~G Hafemann, Luiz~S Oliveira, Ismail~Ben Ayed, Robert Sabourin, and Eric Granger.
\newblock Decoupling direction and norm for efficient gradient-based l2 adversarial attacks and defenses.
\newblock In \emph{IEEE/CVF Conference on Computer Vision and Pattern Recognition}, pages 4322--4330, 2019.

\bibitem[Sehwag et~al.(2021)Sehwag, Mahloujifar, Handina, Dai, Xiang, Chiang, and Mittal]{sehwag2021robust}
Vikash Sehwag, Saeed Mahloujifar, Tinashe Handina, Sihui Dai, Chong Xiang, Mung Chiang, and Prateek Mittal.
\newblock Robust learning meets generative models: Can proxy distributions improve adversarial robustness?
\newblock \emph{arXiv preprint arXiv:2104.09425}, 2021.

\bibitem[Selvaraju et~al.(2017)Selvaraju, Cogswell, Das, Vedantam, Parikh, and Batra]{Selvaraju2017GradCAM}
Ramprasaath~R. Selvaraju, Michael Cogswell, Abhishek Das, Ramakrishna Vedantam, Devi Parikh, and Dhruv Batra.
\newblock {Grad-CAM: V}isual explanations from deep networks via gradient-based localization.
\newblock In \emph{IEEE International Conference on Computer Vision}, pages 618--626, 2017.

\bibitem[Shafahi et~al.(2019)Shafahi, Najibi, Ghiasi, Xu, Dickerson, Studer, et~al.]{Shafahi2019Adversarial}
Ali Shafahi, Mahyar Najibi, Amin Ghiasi, Zheng Xu, John~P. Dickerson, Christoph Studer, et~al.
\newblock Adversarial training for free!
\newblock In \emph{Advances in Neural Information Processing Systems}, pages 3353--3364, 2019.

\bibitem[Shen et~al.(2023)Shen, Amara, Li, Meyer, Gross, and Clark]{Shen2023Fast}
Lulan Shen, Ibtihel Amara, Ruofeng Li, Brett Meyer, Warren Gross, and James~J. Clark.
\newblock Fast fine-tuning using curriculum domain adaptation.
\newblock In \emph{Conference on Robots and Vision}, 2023.

\bibitem[Sun and Saenko(2016)]{sun2016deep}
Baochen Sun and Kate Saenko.
\newblock {Deep CORAL}: {C}orrelation alignment for deep domain adaptation.
\newblock In \emph{European Conference on Computer Vision Workshop}, 2016.

\bibitem[Tan and Le(2021)]{Tan2021EfficientNetV2}
Mingxing Tan and Quoc~V. Le.
\newblock {EfficientNetV2: S}maller models and faster training.
\newblock In \emph{International Conference on Machine Learning}, pages 10096--10106. {PMLR}, 2021.

\bibitem[Wang et~al.(2021)Wang, Chen, and Zhu]{wang2021survey}
Xin Wang, Yudong Chen, and Wenwu Zhu.
\newblock A survey on curriculum learning.
\newblock \emph{IEEE Transactions on Pattern Analysis and Machine Intelligence}, 44\penalty0 (9):\penalty0 4555--4576, 2021.

\bibitem[{Yosinski} et~al.(2014){Yosinski}, {Clune}, {Bengio}, and {Lipson}]{yosinski2014transferable}
Jason {Yosinski}, Jeff {Clune}, Yoshua {Bengio}, and Hod {Lipson}.
\newblock How transferable are features in deep neural networks?
\newblock \emph{Advances in Neural Information Processing Systems}, 27, 2014.

\bibitem[Zhang(2021)]{zhang2021survey}
Youshan Zhang.
\newblock A survey of unsupervised domain adaptation for visual recognition.
\newblock \emph{arXiv preprint arXiv:2112.06745}, 2021.

\end{thebibliography}
}


\end{document}